\documentclass[10pt,twocolumn,letterpaper]{article}
\usepackage[accsupp]{axessibility}  

\usepackage[final]{cvpr}

\usepackage{graphicx}
\usepackage{amsmath}
\usepackage{amssymb}
\usepackage{booktabs}
\usepackage{times}
\usepackage{epsfig}
\usepackage{amsmath}
\usepackage{tabularx}
\usepackage{multirow}
\usepackage{blindtext}
\usepackage{color}

\usepackage[table]{xcolor}

\usepackage{algorithm,algpseudocode}

\usepackage{pifont}
\newcommand{\cmark}{\ding{51}}%
\newcommand{\xmark}{\ding{55}}%
\algnewcommand\Input{\item[\textbf{Input:}]}%
\algnewcommand\Output{\item[\textbf{Output:}]}%

\newcommand{\mm}{\textcolor[rgb]{0.0,0.0,0.0}}  

\newcommand{\ac}{\textcolor[rgb]{1,0,1}} 

\newcommand{\expp}{\textcolor[rgb]{0,0.5,0}} 

\usepackage[pagebackref,breaklinks,colorlinks]{hyperref}

\usepackage[capitalize]{cleveref}
\crefname{section}{Sec.}{Secs.}
\Crefname{section}{Section}{Sections}
\Crefname{table}{Table}{Tables}
\crefname{table}{Tab.}{Tabs.}

\def\cvprPaperID{84} 

\begin{document}

\title{Auxiliary Learning for Self-Supervised Video Representation via Similarity-based Knowledge Distillation}



\author{Amirhossein Dadashzadeh \\
University of Bristol\\
{\tt\small a.dadashzadeh@bristol.ac.uk}
\and
Alan Whone\\
University of Bristol\\
{\tt\small alan.whone@bristol.ac.uk}

\and
Majid Mirmehdi\\
University of Bristol\\
{\tt\small majid@cs.bris.ac.uk}

}

\maketitle

\begin{abstract}


Despite the outstanding success of self-supervised pretraining methods for video representation learning, they generalise poorly when the unlabeled dataset for pretraining is small or the domain difference between unlabelled data in source task (pretraining) and labeled data in target task (finetuning) is significant. 
To mitigate these issues, we propose a novel approach to complement  self-supervised pretraining via an auxiliary pretraining phase, based on knowledge similarity distillation, auxSKD, 
for better generalisation with a significantly smaller amount of video data, e.g. Kinetics-100 rather than Kinetics-400. Our method deploys a teacher network that iteratively distils its knowledge to the student model by capturing the similarity information between segments of  unlabelled video data. The student model \mm{meanwhile} solves a pretext task by exploiting this prior knowledge. We also introduce a novel pretext task, Video Segment Pace Prediction or VSPP, which requires our model to predict the playback speed of a randomly selected segment of the input video to provide more reliable self-supervised representations. Our experimental results show  superior results to the state of the art on both UCF101 and HMDB51 datasets \mm{when pretraining on K100 in apple-to-apple comparisons.} Additionally, we show that our auxiliary pretraining, auxSKD, when added as an extra pretraining phase to recent \mm{state of the art} self-supervised methods \mm{(i.e. VCOP, VideoPace, and RSPNet),} improves their results on UCF101 and HMDB51. Our code \mm{is available at https://github.com/Plrbear/auxSKD.}
\end{abstract}

\section{Introduction}
\begin{figure}[t]

\centerline{\includegraphics[scale=0.66]{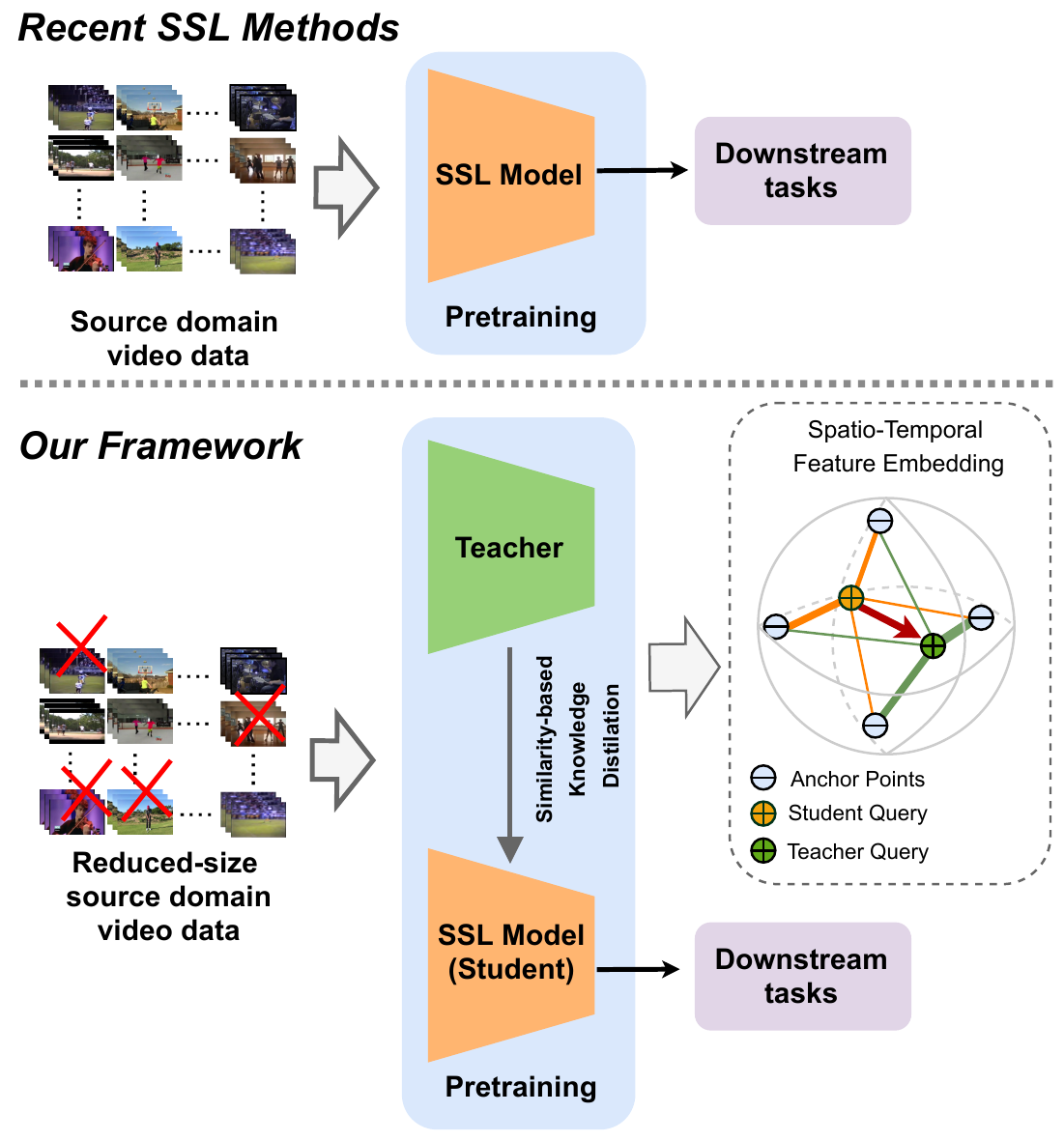}}
\caption{{High-level overview of of our framework and recent SSL methods --  while recent methods encourage their model to solve a pretext task from scratch, our SSL model benefits from an implicit similarity-based knowledge, distilled by a teacher model, before solving the pretext task. However, the question that we pose in this paper is: can we use an implicit knowledge of this type to improve the generalization ability of self-supervised approaches?} }
\label{fig:intro}  
\end{figure}
Self-supervised learning (SSL) methods have taken great strides recently in acquiring high-level semantic visual representations from unlabelled data, eliminating the cost of annotating large-scale datasets \cite{fernando2017self,komodakis2018unsupervised, he2020momentum, wang2020self, chen2021rspnet}.  The most common approach in SSL is to use large-scale unlabelled data, such as Kinetics-400 \cite{kay2017kinetics}, for pretraining network parameters, followed by finetuning on a downstream task with a limited amount of {labelled} data \cite{fernando2017self,grill2020bootstrap:, he2020momentum,wang2020self,chen2021rspnet}. Despite {the reduction in} manual labelling, the  performance of SSL methods is leveraged on huge unlabelled datasets, which demands high computational and memory costs, especially for large-scale video datasets. For example, it takes around two weeks to train MoCo \cite{he2020momentum} on Kinetics-400 for 300 epochs with two Nvidia RTX 2080TI GPUs. In fact, such computational costs place many {state of the art (SotA)} SSL approaches only in the realms of huge corporations who have such powerful resources \cite{chen2021rspnet,grill2020bootstrap:,pan2021videomoco}, and this further becomes a subject of ethical fairness as well as carbon emission footprints \cite{strubell2020energy}.

{A few recent works have addressed the issue of efficient pretraining for {\it image-based} tasks \cite{liu2019self, henaff2021efficient, reed2021self}, but there is only Lin et al. \cite{lin2021self}'s work for {\it video-based} tasks which improves the generalization performance of a contrastive learning-based method \cite{tian2020contrastive} under a meta-learning paradigm. However, their method is {not} suited to most of the SotA works that use transformation-based pretext tasks \cite{fernando2017self,luo2020video, wang2020self, jenni2020video, wang2021unsupervised}.}

{Our motivation is to develop a {\it task-agnostic pretraining process that alleviates the dependency on large-scale datasets} for self-supervised video representation learning, while ensuring the model generalises well and still contains rich information. To achieve this, we propose an auxiliary {pretraining} stage, based on knowledge distillation (KD), which trains on a reduced version of the source dataset, provides implicit knowledge for the {primary} pretraining stage {with the same reduced-size source dataset}, and boosts generalization for downstream video representation learning tasks.}
Figure \ref{fig:intro} illustrates the difference
between our framework and existing SSL methods, such as \cite{fernando2017self,luo2020video, wang2020self, jenni2020video, chen2021rspnet}.
{We employ a slowly progressing teacher model to iteratively distill knowledge to the student, our self-supervised model, by evaluating the similarity information of an augmented view of a query {video clip} to a large {queue} of random {clips} as anchors and transferring that information to the student. To the best of our knowledge, this is the first time such Similarity-based {Knowledge Distilation} (SKD) has been used in video-based self-supervised learning, while recently, SKD was adopted in image-based applications, for example for contrastive learning  \cite{tian2019contrastive:,tejankar2021isd} or {model compression} \cite{abbasi2020compress,fang2021seed:}. {To refer to this aspect of our work, we use auxSKD.}
To support the operation of the proposed approach on temporal features in the video domain, 
we  apply temporal augmentations, in addition to spatial augmentations, to generate different transformed versions of a query video.} 
Such temporal transformations are the same as the pretext transformations used in the primary pretraining stage. Their application at this stage allows our teacher to impart knowledge which matters most in the primary pretraining stage.  


{Also in this paper}, we propose a new pretext task for video representation learning, namely Video Segment Pace Prediction (VSPP). While recent video playback rate prediction methods randomly sample training clips at different paces {or speeds} \cite{benaim2020speednet, wang2020self}, we {sample training clips where only a randomly selected segment of the video has a randomly selected speed and the other segments of the  video retain their natural pace.} VSPP then requires the {learner model} to predict the playback speed of this {randomly selected segment {and its temporal location in} the input training video}. {We advocate that by solving this pretext task, our model can strengthen its awareness of the natural pace of the clip and deal with the imprecise video speed labeling problem \cite{chen2021rspnet}.}

\mm{In our experiments, we show that our results, {\it based on Kinetics-100 pretraining as an example of a reduced-size dataset,} rather than the commonly used Kinectics-400, beat VCOP \cite{xu2019self} ($\uparrow$4.7\% on UCF101 and $\uparrow$7.8\% on HMDB51), VideoPace \cite{wang2020self} ($\uparrow$5.4\% on UCF101 and $\uparrow$9.5\% on HMDB51) and RSPNet \cite{chen2021rspnet} ($\uparrow$2.7\% on UCF101 and $\uparrow$2.3\% on HMDB51) in like-for-like comparisons}. 


Our key contributions can be summarised as follows:

(i)  we propose an auxiliary pretraining stage for self-supervised video learning to alleviate the dependency on large-scale source datasets, e.g. to allow using Kinetics-100 instead of Kinetics-400, 
(ii) we extend similarity-based knowledge distillation to the task of self-supervised video representation learning {which has not been shown before}, 
(iii) we show that our approach can benefit other pretext tasks for self-supervised pretraining that involve video transformations (e.g. VCOP \cite{xu2019self}, VideoPace \cite{wang2020self} and RSPNet \cite{chen2021rspnet}), 
(iv) we propose a simple, yet effective and novel pretext task which is more commensurate with video motion {events} than existing video playback prediction tasks, 
(v) we achieve SotA results  \mm{\it when pretraining with Kinetics-100,} and evaluate the performance of different components of our proposed method through ablation experiments.  

\begin{figure*}[h]
\centerline{\includegraphics[scale=0.25]{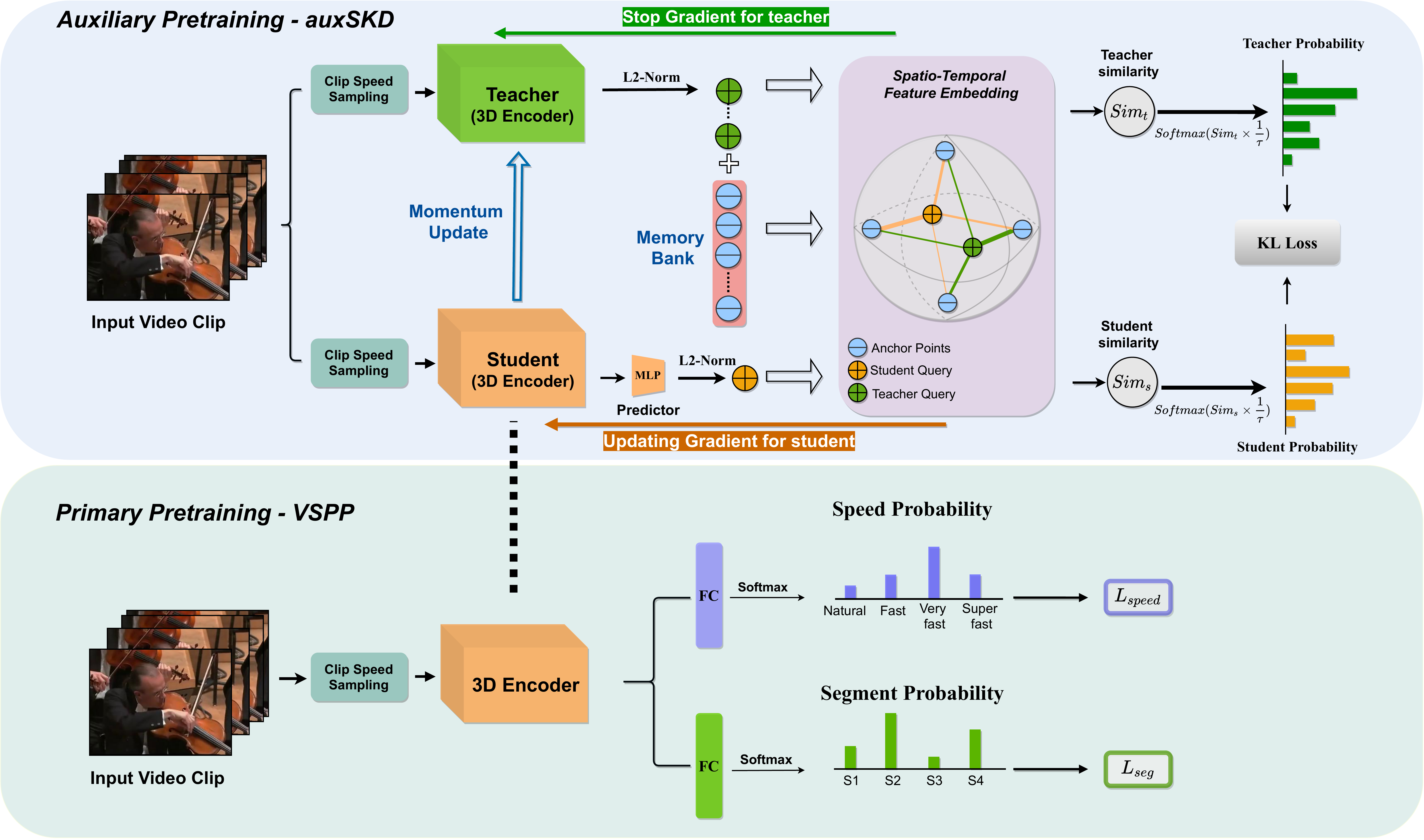}}
\caption{{The self-supervised learning pretext training scheme is supported by an Auxiliary Pretraining task ({auxSKD} - see top region) that provides a similarity knowledge distillation process via a teacher-student configuration. In this configuration, both the teacher and the student 3D encoders are initialized and trained from scratch. Our teacher encoder is updated using momentum as a moving-average of the student weights. We train the student via gradient update by minimizing the KL divergence between the two probabilities from the teacher and the student for a transformed version of input video $v$, computing its similarity over anchor points. Note that in each iteration our encoders randomly take a different transformed input via {our clip speed sampling process (see section \ref{sec:VSPP}).}
In the primary pretraining task (see bottom region), the student is ready to solve our VSPP task on input clips with segments that include changed pace. }}

\label{fig:network}  
\end{figure*}

\section{Related Work}
{In this section, we pay attention to three of the most prominent areas within the main scope of our work, with focus on the more recent, SotA approaches. }



\textbf{Auxiliary Learning --} {To assist a primary task  to generalise better to unseen data, training through auxiliary learning is an effective approach \cite{liu2019self,shi2020auxiliary, navon2020auxiliary} and has been applied alongside a wide range of techniques, such as transfer learning \cite{tzeng2015simultaneous}, reinforcement learning \cite{lin2019adaptive}, semi-supervised learning \cite{zhai2019s4l}, and knowledge distillation \cite{xu2020knowledge}. {In a recent,  work, Xu et. al \cite{xu2019self} treat a self-supervised learning method as an auxiliary task for knowledge distillation, pushing the student to mimic the teacher on both classification output and auxiliary self-supervision pretext task output. They show that the auxiliary learning step regularizes the student to generalize better on few-shot and noisy-label scenarios.} 
{Here, we show that the similarity information between embedded feature points can be used as implicit knowledge for self-supervised pretraining to learn more generalised representations through pretext tasks. To capture this similarity information, we employ a variation of knowledge distillation, called similarity-based knowledge distillation \cite{tian2019contrastive:, tung2019similarity}, as an auxiliary task which can be applied to any self-supervised pretraining method.} }



\textbf{Similarity-based Knowledge Distilation --}
{Knowledge distillation  \cite{buc2006,hinton2015distilling} establishes a framework  for improving the performance of a lightweight student model, guided by a larger and better performing teacher model which distills the (dark) knowledge it learns to its student \cite{hinton2015distilling, zagoruyko2016paying, yim2017gift, noroozi2018boosting}.}

{Similarity-based Knowledge Distillation (SKD) methods \cite{tian2019contrastive:, tung2019similarity, peng2019correlation, park2019relational, fang2021seed:, abbasi2020compress, tejankar2021isd} train a student to mimic the similarity score distribution inferred by the teacher over data samples. Most early works in SKD use a supervised loss during distillation \cite{tung2019similarity, peng2019correlation, park2019relational}, but there are a number of works that combine SKD with contrastive SSL learning to achieve SSL model compression and performance gains \cite{abbasi2020compress, fang2021seed:}. While these works rely on a pre-trained frozen teacher model, Tejankar et al. \cite{tejankar2021isd} propose an iterative SKD regime where the teacher model continues to learn similarity score distributions during training.} Our auxSKD architecture  is similar to \cite{tejankar2021isd}, but our objectives are different, (i) we expand SKD to extract representations from video data, instead of images, and (ii) we use distilled similarity representations as auxiliary knowledge for different self-supervised pretraining methods, instead of directly applying them  on downstream tasks.

\textbf{Video Playback Speed Perception -- } 
{Creating an efficient self-supervised pretext task to model motion and appearance for video SSL is significantly more difficult than for static images \cite{chen2021rspnet}. Recently, estimating video playback speed has attracted much interest as a highly effective way to encourage the model to learn features (of moving objects) in videos \cite{epstein2020oops, benaim2020speednet, wang2020self, jenni2020video, yao2020video, chen2021rspnet, huang2021ascnet}. For example, Epstein et al. \cite{epstein2020oops} design a method to predict normal video speed to detect an unintentional event in the video. SpeedNet \cite{benaim2020speednet} determines whether a given video clip is being played at normal or twice its original speed, while VideoPace \cite{wang2020self}  predicts the specific speed of each video clip which is randomly sampled at a different frame rate. One of the main limitations in considering  playback speed alone is that video clips with different speed labels might appear similar to each other, e.g. when different athletes might perform the same sporting action at different speeds.}

{To avoid the dependence on imprecise speed labels, Chen et al. \cite{chen2021rspnet} introduce RSPNet to predict relative speed between two video clips to better learn motion features. They use a triplet loss to minimize the distance between two clips of the same video at the same playback speed and maximize the distance between two clips of the same video at different playback speeds. Also in ASCNet \cite{huang2021ascnet}, Hunag et al. focus on speed similarity and propose a consistent speed perception task which aims to minimize the distance between two clips from two different videos with the same playback speed.} 

{In VSPP, we propose a simple, yet effective video sampling strategy which does not rely on comparing video clips at different speed rates, since we embed a speed rate change within each clip. This emphasises focus on motion and  abstracts the model from appearance features for which RSPNet and ASCNet require a whole additional processing pipeline. It also allows our model to converge much faster, e.g. {after 20 epochs} compared to 200 for RSPNet and ASCNet.}

\section{Proposed Approach}
Our goal is {to reduce the pretext training computational burden by developing  an auxiliary pretraining phase that assists the primary pretext task to learn as efficient generalised self-supervised video representation as possible on a reduced-size source dataset. To achieve this, we take inspiration from Similarity-based Knowledge Distillation which is used in recent works \cite{tian2019contrastive:,tejankar2021isd,abbasi2020compress,fang2021seed:}.} We illustrate our full self-supervised pretraining framework in Figure \ref{fig:network}.  



\subsection{Auxiliary Learning via {auxSKD}} \label{sec:auxSKD}
{Our auxiliary learning framework consists of a teacher $\mathcal{T}$ and a student $\mathcal{S}$ with the same architecture followed by a {fully-connected layer, as the projection,} to map the representations into a lower dimension space. We follow BYOL \cite{grill2020bootstrap:} and use a MLP predictor layer on top of the student model. This makes the {teacher-student} architecture asymmetric to prevent collapsed solutions. We randomly initialize  both models from scratch equally.  The student model and its predictor layer are updated by back-propagation while momentum update \cite{he2020momentum}  is applied in the teacher model to be a running average of the student.}

At each iteration our pretext task transformation VSPP is applied twice to a raw video instance $v$ to generate two video clips $v^*_1$ and $v^*_2$ independently, 
with the goal of maximizing their similarity in our teacher-student framework. Then, given  feature encodings $({\mathcal{T}}(v^*_1),\mathcal{S}(v^*_2))$ and 
predictor function $\text{MLP}(.)$ for $\mathcal{S}$, we perform $L_2$ normalization such 
that $z^{\mathcal{T}} = {\mathcal{T}}(v^*_1)/\|{\mathcal{T}(v^*_1)\|}_2$ and 
$z^\mathcal{S} = {\text{MLP}(\mathcal{S}}(v^*_2))/\|{\text{MLP}(S(v^*_2))\|}_2$.


 Similar to \cite{he2020momentum,tejankar2021isd}, we consider a memory bank of $H$ feature vectors (or anchors) $x_i^\mathcal{T}=[x_{1}^{\mathcal{T}},...,x_{H}^{\mathcal{T}}]$ obtained from the teacher model under a simple FIFO strategy. Specifically, at each iteration, we enqueue the feature vectors of the current batch extracted from the teacher model and dequeue the earliest instances. Next, we calculate the similarity of the teacher’s embedding $z^\mathcal{T}$  to all feature vectors in the memory bank and {apply Softmax to obtain a probability distribution,}

\begin{equation}
\label{eq:eq1}
    p_{i}^\mathcal{T} = -\log \frac{\text{exp}(\text{sim}(z^\mathcal{T},x_i^\mathcal{T} )/\gamma^\mathcal{T})}{\sum_{j=1}^{H} = \text{exp}(\text{sim}(z^\mathcal{T},x_j^\mathcal{T} )/\gamma^\mathcal{T})} ~,
\end{equation}
{where $p^\mathcal{T}_{i}=[p_{1}^{\mathcal{T}},...,p_{H}^{\mathcal{T}}]$ 
is the probability of teacher query $z^\mathcal{T}$ for the $i$-th anchor point,
$\text{sim}(.,.)$  measures the similarity} between $L_2$ vectors, and $\gamma^\mathcal{T}$ is the temperature value for the teacher's model. 

Similarly, we calculate the student similarity {distribution $p^\mathcal{S}_{i}=[p_{1}^{S},...,p_{H}^{S}]$ over anchor points}, with
\begin{equation}
\label{eq:eq2}
    p_{i}^\mathcal{S} = -\log \frac{\text{exp}(\text{sim}(z^\mathcal{S},x_i^\mathcal{T} )/\gamma^\mathcal{S})}{\sum_{j=1}^{H} = \text{exp}(\text{sim}(z^\mathcal{S},x_j^\mathcal{T} )/\gamma^\mathcal{S})} ~.
\end{equation}
Here $\gamma^\mathcal{S}$ is the temperature value for the student's model.
Finally, the loss is measured by the Kullback–Leibler (KL) divergence as 
\begin{equation}
\label{eq:KL-Loss}
    \mathcal{L}(\mathcal{T},\mathcal{S}) = \sum _i \text{KL}(p^\mathcal{T}_i~ \| ~ p^\mathcal{S}_i) ~.
\end{equation}
Note that during training, 
the teacher network's weights are initialised randomly and then they evolve gradually as a running
average of the student using momentum with the update rule   
$\theta_\mathcal{T} \gets m \theta_\mathcal{T} + (1-m)\theta_\mathcal{S}$, 
where $m \in [0,1) $ is the momentum hyperparameter to ensure smoothness and stability, and $\theta_\mathcal{T}$ and $\theta_\mathcal{S}$ are the teacher and student model parameters respectively.
Pseudo-code for our {auxSKD} training is provided in Algorithm \ref{alg:auxSKD}. 

\begin{algorithm}[t]
\caption{Training {auxSKD}}  \label{alg:auxSKD}
\begin{algorithmic}[1]

    \Input
  {
    Teacher model $\mathcal{T}(., \theta_\mathcal{T})$ and student model $\mathcal{S}(., \theta_\mathcal{S})$, videos $V=\{v_i\}_{i=1}^{N}$, and {memory bank} $\{x_i^\mathcal{T}\}_{i=1}^H$. }
    \Output{Trained student model weights $\theta_\mathcal{S}$.}
    \State {{Randomly initialize} $\mathcal{T}(., \theta_\mathcal{T})$ and $\mathcal{S}(., \theta_\mathcal{S})$. }
\While{not max epoch}
\State \parbox[t]{\dimexpr\linewidth-\algorithmicindent}{%
Randomly sample a video $v$ from $V$.}
\State \parbox[t]{\dimexpr\linewidth-\algorithmicindent}{%
Sample two clips {$v^*_1$} and {$v^*_2$} from $v$ using VSPP video transformation (Section \ref{sec:VSPP}).}
\State \parbox[t]{\dimexpr\linewidth-\algorithmicindent}{%
Compute student query features $z^\mathcal{S}$ from clip {$v^*_1$} using student model $\mathcal{S}(., \theta_\mathcal{S})$.}
\State \parbox[t]{\dimexpr\linewidth-\algorithmicindent}{%
Update teacher {parameters using momentum: 
 $\theta_\mathcal{T} \gets m \theta_\mathcal{T} + (1-m)\theta_\mathcal{S}$.}}
\State \parbox[t]{\dimexpr\linewidth-\algorithmicindent}{%
Compute teacher query features $z^\mathcal{T}$ from  clip $v^*_2$ using teacher model $\mathcal{T}(., \theta_\mathcal{T})$.} 
\State \parbox[t]{\dimexpr\linewidth-\algorithmicindent}{%
Calculate $p_i^\mathcal{T}$ and $p_i^\mathcal{S}$ using Eq. \ref{eq:eq1} and Eq. \ref{eq:eq2}.} 
\State \parbox[t]{\dimexpr\linewidth-\algorithmicindent}{%
Add the teacher's embedding $z^\mathcal{T}$ into the memory bank $\{x_i^\mathcal{T}\}_{i=1}^H$ and pop-out the earliest sample. }
\State\parbox[t]{\dimexpr\linewidth-\algorithmicindent}{%
Optimize student model using KL divergence loss, Eq. \ref{eq:KL-Loss}. }

\EndWhile
\end{algorithmic}
\end{algorithm}

\subsection{Primary Pretext task learning via VSPP}
\label{sec:VSPP}
A SSL pretext task encourages the neural network to learn a representation from unlabelled data which contains high-level abstractions or semantics.  In the video pace prediction approach of Wang et al. \cite{wang2020self}, each training clip is randomly sampled at a different pace and their pretext task then identifies the pace for each clip. While this is an effective approach, it means each clip  is treated as if its pace is its natural speed.  

We propose that each clip should contain within it one segment where the pace has been (randomly) altered. {Our assumption is similar to \cite{wang2020self,benaim2020speednet,chen2021rspnet} in that the network can only represent the underlying video content through efficient spatiotemporal features if it succeeds in learning the pace reasoning task, however, we build on \cite{wang2020self}'s proposal through a more intricate, yet simple, within-video pace alteration task.} {Our proposed VSPP pretext task requires our model to temporally explore a video clip and predict the index {and speed} of a segment within a clip which is sampled at a different speed rate.

Given a video clip {$v_i$} comprising $N$ frames,  we generate video {$v^*_i=\{I_0,I_1,...,I_{K-1}\}$} of size $K\!\!<\!\!N$, comprising $Z$ segments, such that $K/Z$ number of frames in segment $\zeta$ are sampled at pace $\lambda$, where both $\zeta$ and $\lambda$ are randomly selected from  $1 \leq \zeta \leq Z$ and $1\leq \lambda \leq Q$ respectively,  and $Q$ is the highest possible speed rate. Note, when ${Z=1}$ the sampling strategy is similar to \cite{wang2020self}. In this work, we select $Z=4$ and $Q=4$ to allow a significantly wide variation of starting locations and {sudden speed rate changes} to provide more precise self-supervision signals. {Specifically,} our approach results in a change of speed rate in only one segment of the clip while the rest of the clip (before and after) retains its natural rate {(see Figure \ref{fig:vidsample}). {This strategy allows the network to better find the difference between natural speed (changes which happen gradually) and altered speed (changes which happen suddenly) in a clip, alleviating imprecise video speed labeling issues \cite{chen2021rspnet}.} }
To have a random speed rate $\lambda$ for the $\zeta^{th}$ segment,  beginning at frame $I_b$ and ending at frame $I_e$, then
\begin{flalign}
   { I_b = f_r + ((\zeta-1)*\frac{K}{Z}) + (\lambda -1) } , \nonumber\\
  I_e = I_b + \lambda*(\frac{K}{Z}-1 )~,
\end{flalign}
where $f_r$ is the $r^{th}$ frame of the original video $v_i$ which is randomly selected during sampling to generate a more diverse video clip $v_i^*$ at each iteration.  \\

\begin{figure}[t]
\centerline{\includegraphics[scale=0.43]{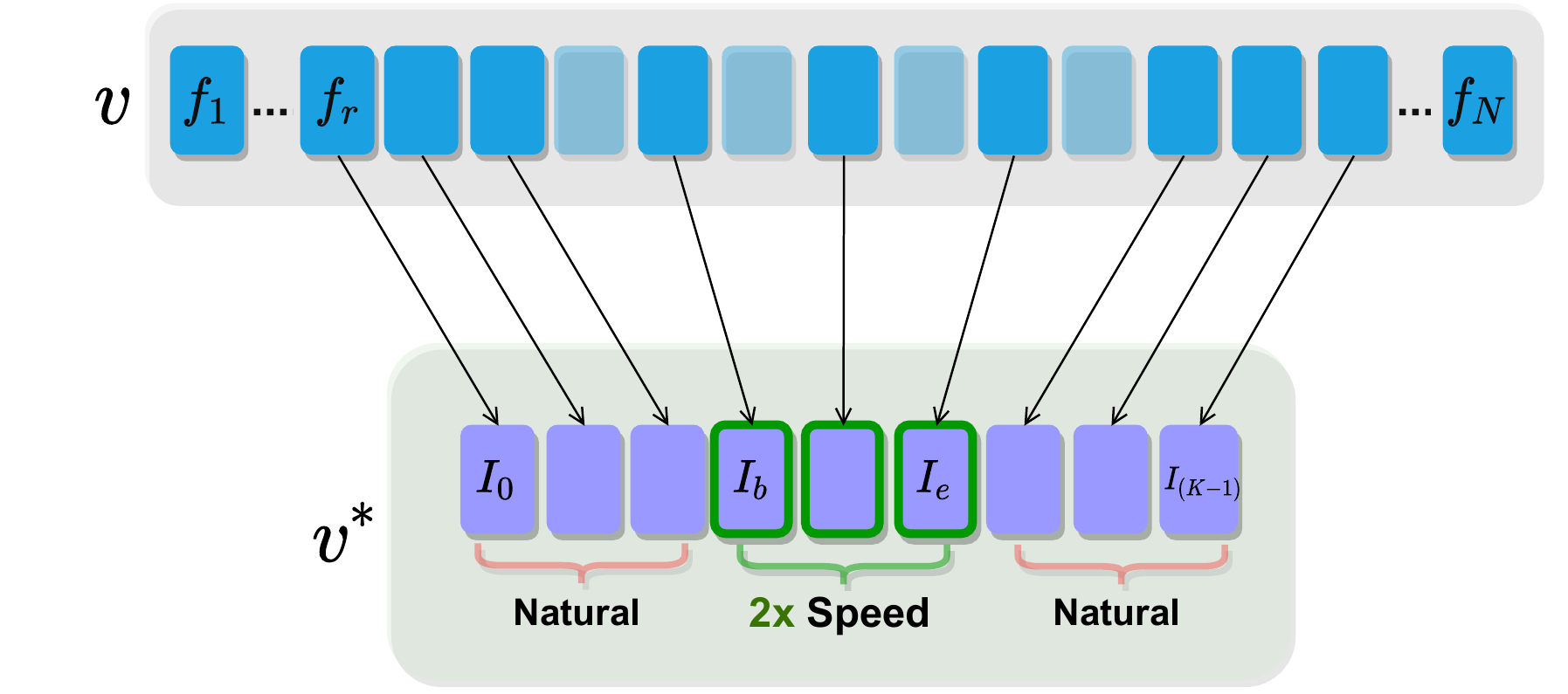}}
\caption{{Changing the natural speed of one random segment of a video clip for the pretraining stage - the VSPP pretext task {learns where in $v^*$ this occurs and at what speed change. In this example $\lambda=2$ and $\zeta=2$.}}}
\label{fig:vidsample}  
\end{figure}

{\bf Summary of our overall method}:
{Given a 3D encoder, such as an R(2+1)D or R3D-18, and a video dataset $V=\{v_i\}_{i=1}^{N}$, we perform our auxiliary learning stage using our flavour of SKD (see Section \ref{sec:auxSKD} and Algorithm \ref{alg:auxSKD}) based on a KL loss (Eq. \ref{eq:KL-Loss}). 
Following this auxiliary pretraining stage, the student model enters the primary pretraining stage {to solve our VSPP pretext task through two simultaneous sub-tasks: i) predicting the speed rate {$\lambda$} in the $\zeta^{th}$ segment of $v^*$, ii) predicting the temporal location of the segment in $v^*$ which is sampled at a different speed, i.e. predicting index $\zeta$. Then by jointly optimizing these two tasks, the final self-supervised loss is defined as: }
\begin{equation}
     \mathcal{L} = \alpha\mathcal{L}_{speed} + \beta \mathcal{L}_{seg},
\end{equation}
{where $\alpha$ and $\beta$ are balancing weights (empirically found to work best in our experiments when $\alpha=\beta=1$). $\mathcal{L}_{speed}$ and $\mathcal{L}_{seg}$ are cross-entropy losses.}





\section{Experiments}

\subsection{Datasets}

 We conducted our experiments on four datasets, {two for pretraining, Kinetics-400 \cite{kay2017kinetics} (K-400) and Kinetics-100 \cite{chen2021rspnet} (K-100), and two for downstream action recognition, UCF101 \cite{soomro2012ucf101}, HMDB51 \cite{kuehne2011hmdb}.} 

{\bf Kinetics-400} {is a huge dataset for action
recognition collected from Youtube, which contains 400 human action classes and around 240K videos. Each video lasts about 10 seconds.}

{\bf{Kinetics-100}} comprises 100 classes and around 33K videos. We use K-400 and K-100 as pretraining datasets to validate our proposed approach's performance on a reduced-size dataset and promote less dependency on large-scale datasets for self-supervised representation learning.

{\bf UCF101} contains 101 action categories with a total of 13320 videos. It is divided into three training/testing splits and we follow prior works \cite{wang2020self, chen2021rspnet} to use training split 1 for self-supervised pre-training and train/test split 1 for fine-tuning and evaluation. 

{\bf HMDB51} contains 6,849 video clips of 51 action classes -- a relatively small dataset {compared to UCF101 and Kinetics}. It is divided into three splits and we use split 1 for our downstream task evaluation, similar to \cite{wang2020self, chen2021rspnet}. 

{\subsection{Implementation Details}}

{\bf Backbone Networks} -- We choose two different backbones R(2+1)D \cite{tran2018closer} and R3D-18 \cite{hara2018can}, as our 3D encoder, which have been widely used in recent SotA self-supervised video representation learning methods \cite{pan2021videomoco,chen2021rspnet, huang2021ascnet}. 

{{\bf Default Settings} -- We run all experiments under PyTorch on two GeForce RTX 2080Ti GPUs with a batch size of 30. We use SGD as our optimizer with momentum of 0.9 and weight decay of 5e-4.}

{\bf Pretraining Stages} -- {Following \cite{wang2020self}}, for both auxiliary and primary stages, we pretrain our models {for 20 epochs} with an initial learning rate  of $1 \times 10^{-3}$. The learning rate is decreased {by} 1/10 
every 6 epochs. For data augmentation, we randomly crop the video clip to 112 × 112 and then apply horizontal flip and color jittering to each video frame. Following \cite{alwassel2020self}, for UCF101 we apply (10x more iterations at) 90K iterations per epoch for temporal jittering. 
In our auxiliary pretraining stage, we use a predictor head for the student encoder comprising a 3-layer MLP with hidden dimension 1024, and output embedding dimension 128. We do not use a predictor for the teacher and only set its output dimension ({projection head}) to 128. {We follow MoCo \cite{pan2021videomoco} and set the size of the memory bank to 16384 and set the momentum value of the encoder update to 0.999}. We also use the same temperature for both teacher and student model at 0.02.
To use the student encoder for the primary stage, the weights of the convolutional layers are retained after auxiliary pretraining and we {drop} the {projection layer and predictor to replace them} with two randomly initialized FC layers corresponding to the 
segment speed and index outcomes of our VSPP pretext task  (see Figure \ref{fig:network}). We select our parameters empirically. 

{{\bf Fine-tuning} -- During fine-tuning, we transfer the weights of the convolutional layers to the human action recognition downstream task, while the last FC layer is randomly initialized. 
We fine-tune the network on UCF101 and HMDB51 for 25 epoches with labelled videos and apply cross-entropy loss. We use the same data augmentation  and training strategy as the pretraining stage except for the initial learning rate which is set to $3 \times 10^{-3}$, {similar to \cite{wang2020self}.}

{{\bf Evaluation Settings} -- We follow the common evaluation protocols on video representation learning \cite{wang2020self, huang2021ascnet} to assess the performance of our proposed approach. For action recognition, we sample 10 clips uniformly from each video in the test sets of UCF-101 and HMDB-51. Then for each clip, we only simply apply the center-crop. To find the final prediction, we average the Softmax probabilities of all 10 clips from the video.}

\subsection{Evaluations on UCF101 and HMDB51}
{{\bf Comparison on K400 pretraining}} --  \mm{For completeness sake, and to illustrate how our proposed method fares when pretrained on K400,  we present comparative results in Table \ref{tab:compK400-100}} for top-1 accuracy on both UCF-101 and HMDB-51 datasets, along with the pretraining settings for all methods,  i.e. backbone architecture, input size, pretraining dataset, {and number of epochs}. {In Rows 1-4, we show a mix of methods that operate on temporal manipulations at different input sizes and on different backbones for reference. Rows 5-10 allow more like-for-like comparisons of recent, popular works in SSL video representation learning based on K-400 pretraining, R3D-18 backbone and almost consistent image sizes across the techniques.  ASCNet achieves the most superior results with a combined appearance and speed manipulation approach.}
{In Rows {11-15}, where pretraining is on K-400 on the  R(2+1)D architecture, RSPNet and VideoMoCo come 1st and 2nd-best alternatively on the two test datasets, while our approach exceeds CEP on both.}


{\bf Comparison on K100 pretraining} -- {The results on Rows 16-{29} of Table \ref{tab:compK400-100}} represent  the essence of our contributions, in that we aim to reduce the dependence of SSL methods on large pretraining datasets, for example by replacing K-400 with K-100 for pretraining. We apply our auxSKD stage to {two other} transformation-based pretext tasks, i.e. {VCOP}, VideoPace,  and also to one contrastive task, i.e. RSPNet, to exhibit the flexibility of our method. {For VCOP and VideoPace, we train their auxSKD with video clips sampled based on VCOP and VideoPace's own sampling strategies, as proposed in \cite{wang2020self} and \cite{xu2019self} respectively. } To integrate auxSKD into  the RSPNet framework, we train it with the video transformation proposed in \cite{chen2021rspnet} and then transfer all the convolutional layer weights to its query encoder and initialize the projection head {and key encoder} randomly from scratch.

\mm{Rows 16-22 relate to the networks with a R(2+1)D {backbone}. VCOP+auxSKD improves on VCOP by $\uparrow$1.2\% and $\uparrow$0.4\%  on UCF101 and HMDB51 respectively, while VideoPace+auxSKD similarly surpasses VideoPace alone by  $\uparrow$2.3\% and $\uparrow$2.4\%. Note these are very close performances to when VideoPace is pretrained on K-400 (cf. Row 11).  RSPNet's performance also improves when our auxiliary SKD is deployed, by  $\uparrow$0.8\% for UCF101 and $\uparrow$1.7\%   for HMDB51.} 

\mm{When the R3D-18 backbone is used, consistent improvements are again observed (see Rows 23-29) for all these methods when auxSKD is added to them. Our proposed method obtains the best performance using the R(2+1)D backbone on both datasets at 76.3\% and 39.6\%. When using R3D-18, it achieves  the best result on UCF101 at 62.9\%  and the 2nd best on HMDB51 at 33.0\%.
Finally, we note that unlike VideoPace \cite{wang2020self}, RSPNet \cite{chen2021rspnet} and ASCNet \cite{huang2021ascnet}  (for which no code has been released at the time of writing), we do not have an appearance stream in our method.}}


\begin{table*}[t]   \label{tab:compK400-100}
\footnotesize	
\centering
\begin{tabular}{rlcccr|ll}
\hline
\multicolumn{6}{c|}{\bf Self-Supervised Methods}                                       & \multicolumn{2}{c}{\bf Top 1 accuracy} \\
{\bf Row}  & {\bf Method}            & {\bf Network}  & {\bf Input Size}                    & {\bf Pretrain }  & {\bf \#epochs} & {\bf UCF101}           & {\bf HMDB51}          \\ \hline
\rowcolor{gray!10} 1 & Shuffle\&Learn \cite{misra2016shuffle} {[ECCV, 2016]} & Alexnet  & 256 $\times$ 256 & UCF101   &      -     & 50.2             & 18.1            \\
\rowcolor{gray!10} 2 & OPN \cite{lee2017unsupervised} {[ICCV, 2017]}              & VGG      & 80 $\times$ 80   & UCF101    &    -      & 59.8             & 23.8            \\
\rowcolor{gray!10} 3 & VCOP \cite{xu2019self} {[CVPR, 2019]}              & C3D  & 112 $\times$ 112 & UCF101      &   -     & 65.6             & 28.4            \\
\rowcolor{gray!10} 4 & SpeedNet \cite{benaim2020speednet} {[CVPR, 2020]}          & I3D      & 224 $\times$ 224 & K-400     &   n/a     & 66.7             & 43.7            \\ 


\hline


\rowcolor{gray!10} 5 & VideoPace \cite{wang2020self} {[ECCV, 2020]}       & R3D-18  & 112 $\times$ 112 & K-400    &     18      & 63.7             & 27.9            \\
\rowcolor{gray!10} 6 & VideoMoCo \cite{pan2021videomoco} {[CVPR, 2021]}     & R3D-18   & 112 $\times$ 112 & K-400    &    200       & 74.1             & \underline{43.6}            \\
\rowcolor{gray!10} 7 & RSPNet \cite{chen2021rspnet} {[AAAI, 2021]}      & R3D-18   & 112 $\times$ 112 & K-400     &   200       & {74.3}             & {41.8}            \\
\rowcolor{gray!10} 8 & ASCNet \cite{huang2021ascnet} [ICCV, 2021]           & R3D-18   & 112 $\times$ 112 & K-400   &     200    & {\bf 80.5}             & {\bf 52.3}            \\
\rowcolor{gray!10} 9 & CEP\cite{yang2020back} {[BMVC, 2021]}         & R3D-18   & 224 $\times$ 224 & K-400      &    50     & \underline{75.9}             & 36.6            \\

\rowcolor{gray!10} 10 & Ours     & R3D-18   & 112 $\times$ 112 & K-400    &   40        & 67.9                & 32.6               \\ 

\hline

\rowcolor{gray!10} 11 & VideoPace \cite{wang2020self} {[ECCV, 2020]}         & R(2+1)D  & 112 $\times$ 112 & K-400  &   18          & 77.1             & 36.6            \\
\rowcolor{gray!10} 12 & VideoMoCo \cite{pan2021videomoco} {[CVPR, 2021]}         & R(2+1)D  & 112$\times$ 112   & K-400  &   200          & \underline{78.7}             & \bf{49.2}            \\
\rowcolor{gray!10} 13 & RSPNet \cite{chen2021rspnet} {[AAAI, 2021]}            & R(2+1)D  & 112 $\times$ 112 & K-400  &    200         & \bf{81.1}             & \underline{44.6}            \\

\rowcolor{gray!10} 14 & CEP\cite{yang2020back} {[BMVC, 2021]}             & R(2+1)D  & 224 $\times$ 224 & K-400   &     50       & 76.7             & 37.6            \\

\rowcolor{gray!10} 15 & {Ours}      & R(2+1)D  & 112 $\times$ 112 & K-400    &    20       & 77.6   & 40.4               \\ \hline




\rowcolor{gray!30} 16 & \mm{VCOP \cite{xu2019self} {[CVPR, 2019]} }        & R(2+1)D  & 112 $\times$ 112 & K-100     & 200    & 71.4    & 32.1            \\

\rowcolor{gray!30} 17 &\mm{VCOP \cite{xu2019self} + auxSKD }        & R(2+1)D  & 112 $\times$ 112 & K-100     & 200    & 72.6    & 32.5            \\

\rowcolor{gray!30} 18 & VideoPace \cite{wang2020self} {[ECCV, 2020]}         & R(2+1)D  & 112 $\times$ 112 & K-100     & 18    & 73.8    & 36.2            \\

\rowcolor{gray!30} 19 & \mm{VideoPace \cite{wang2020self} + auxSKD }        & R(2+1)D  & 112 $\times$ 112 & K-100     & 18    & \underline{76.1}   & 38.6            \\


\rowcolor{gray!30} 20 & \mm{{RSPNet \cite{chen2021rspnet}} {[AAAI, 2021]}}     & R(2+1)D  & 112 $\times$ 112 & K-100 &     200         & {74.7}         & {37.4}       \\  

\rowcolor{gray!30} 21 & \mm{{RSPNet \cite{chen2021rspnet} + auxSKD} }     & R(2+1)D  & 112 $\times$ 112 & K-100 &     200         & {75.5}         & \underline{39.0}       \\

\rowcolor{gray!30} 22 & Ours         & R(2+1)D  & 112 $\times$ 112 & K-100   &    20        & {\bf{76.3}}   & {\bf{39.6}}     \\  \hline

\rowcolor{gray!30} 23 & \mm{VCOP \cite{xu2019self} {[CVPR, 2019]} }            & R3D-18  & 112 $\times$ 112 & K-100  &  200  & {58.2}     & 25.2           \\

\rowcolor{gray!30} 24 & \mm{VCOP \cite{xu2019self} + auxSKD }            & R3D-18  & 112 $\times$ 112 & K-100  &  200  & {60.7}     & 28.4           \\

\rowcolor{gray!30} 25 & \mm{VideoPace \cite{wang2020self} {[ECCV, 2020]}}           & R3D-18  & 112 $\times$ 112 & K-100  &  18  & 57.5     & 23.5           \\

\rowcolor{gray!30} 26 & \mm{VideoPace \cite{wang2020self} + auxSKD }        & R3D-18  & 112 $\times$ 112 & K-100     & 18    & 60.9   & 27.1           \\

\rowcolor{gray!30} 27 & RSPNet \cite{chen2021rspnet} {[AAAI, 2021]}            & R3D-18  & 112 $\times$ 112 & K-100  &   200  & 60.2     & 32.6            \\
\rowcolor{gray!30} 28 & \mm{RSPNet \cite{chen2021rspnet} + auxSKD }  & R3D-18  & 112 $\times$ 112 & K-100 &     200         & \underline{61.9}           & {\bf{33.4}}           \\ 
\rowcolor{gray!30} 29 & Ours  & R3D-18  & 112 $\times$ 112 & K-100    &      20    & \bf{62.9}      & \underline{33.0}      \\ \hline

\end{tabular}
\caption{Comparative performance results on  UCF101 and HMDB51 when pretraining on K400, and most importantly, on the reduced-size  dataset K-100 (shaded region) to emphasise the power of our proposed approach. Note auxSKD refers to our proposed  auxiliary pretraining stage using similarity-based knowledge distillation.}
\label{tab:compK400-100}
\end{table*}

\subsection{Ablation Studies}
{We perform ablations to establish the effectiveness of our auxiliary pretraining process and our VSPP pretext task.}

{{\bf Effectiveness of auxSKD } -- We verify the impact of our auxiliary pretraining stage by showing its gains in performance. In Table \ref{tab:ablateSKD}, we present the results of our proposed method on both R(2+1)D {and R3D-18} backbones, with and without auxiliary pretraining for UCF101, using K-100 and {K-400} for pretraining. It is clear that in each and every case auxSKD causes an increase in performance. 

\mm{{\bf Temperature parameters } -- We studied the effect of changing temperatures of auxSKD for both teacher and student models and report the results in Table~\ref{tab:temp}. Here we use VSPP as the pretext task for the primary stage. }

\begin{table}[t]
\footnotesize
\centering
\begin{tabular}{lcc|c}
\hline
\multirow{2}{*}{\bf Method}    & {\bf Pretrain} & {\bf UCF101}       & {\bf HMDB51}         \\
 &  {\bf Dataset}   & Top-1   & Top-1    \\ \hline
 
\rowcolor{cyan!20} \multicolumn{4}{c}  {Backbone: R(2+1)D}\\  \hline
\rowcolor{gray!30}Ours - auxSKD   & UCF101     &  76.0   &    37.4  \\ 
Ours     & UCF101     & {\bf 77.3} &   {\bf 38.6}  \\ \hline


\rowcolor{gray!30} Ours - auxSKD   & K-100  & 74.0  &  37.3      \\ 
Ours     & K-100     & {\bf 76.3}  &  {\bf 39.6}      \\ \hline

\rowcolor{cyan!20}\multicolumn{4}{c}{Backbone: R3D-18}\\    \hline
\rowcolor{gray!30}Ours - auxSKD   & K-400  & 65.8  &  28.8      \\ 
Ours     & K-400     & {\bf 67.9}    &  {\bf 32.6}      \\ \hline



\rowcolor{gray!30} Ours - auxSKD   & K-100  & 60.8  &  26.3      \\ 
Ours     & K-100     & {\bf 62.9}   &  {\bf 33.0}       \\ \hline

\end{tabular}
\caption{Ablation of the auxiliary pretraining stage auxSKD with our proposed approach.}
\label{tab:ablateSKD}
\end{table}

\begin{table}[h]
\footnotesize
\centering
\begin{tabular}{c|ccccccc}
\hline
$\gamma^\mathcal{T}$      & 0.01 & 0.02          & 0.05 & 0.07 & 0.1  & 0.01 & 0.02 \\ \hline
$\gamma^\mathcal{S}$      & 0.01 & 0.02          & 0.05 & 0.07 & 0.1  & 0.1  & 0.1  \\ \hline
\textbf{UCF101} & 75.0 & \textbf{76.3} & 75.3 & 74.9 & 74.9 & 75.5 & 75.0    \\ \hline
\end{tabular}
\caption{\mm{Effect of changing the temperatures for our method for UCF101 with R(2+1)D backbone. $\gamma^\mathcal{T}$ and $\gamma^\mathcal{S}$ indicate teacher and student temperatures respectively.}}
\label{tab:temp}
\end{table}

\begin{table}[h]
\footnotesize
\centering
\begin{tabular}{cccc}
\hline
{\bf Speed } & {\bf Segment} & {\bf \#Classes } \\
{\bf Prediction}  &  {\bf Prediction} & {\bf \#speed, \#segment} & \multicolumn{1}{l}{\bf UCF101} \\ \hline
\cmark   & -       & {[}$Q=4$ ,    $Z=1${]}   &                    57.5  \\

\cmark   & \cmark       & {[}$Q=4$ ,     $Z=2${]}   &                57.5      \\
\cmark   & \cmark   & {[}$Q=4$ ,     $Z=3${]}        &            59.5  \\ 
\cmark   & \cmark    & {[}$Q=4$ ,     $Z=4${]}    &               \bf{60.8}   \\ 

\cmark  & \xmark    & {[}$Q=4$  ,   $Z=4${]}     &    58.3 \\
\xmark  & \cmark   & {[}$Q=4$   ,     $Z=4${]}  &             59.9     \\ \hline

\end{tabular}
\caption{Ablation of our VSPP pretext task pretrained on K-100 with R3D-18 (no auxSKD stage). We examine the importance of each subtask within VSPP while the number of segments within the clip changes.}
\label{tab:ablateVSPP}
\end{table}

\begin{figure}[h]
\centerline{\includegraphics[scale=0.75]{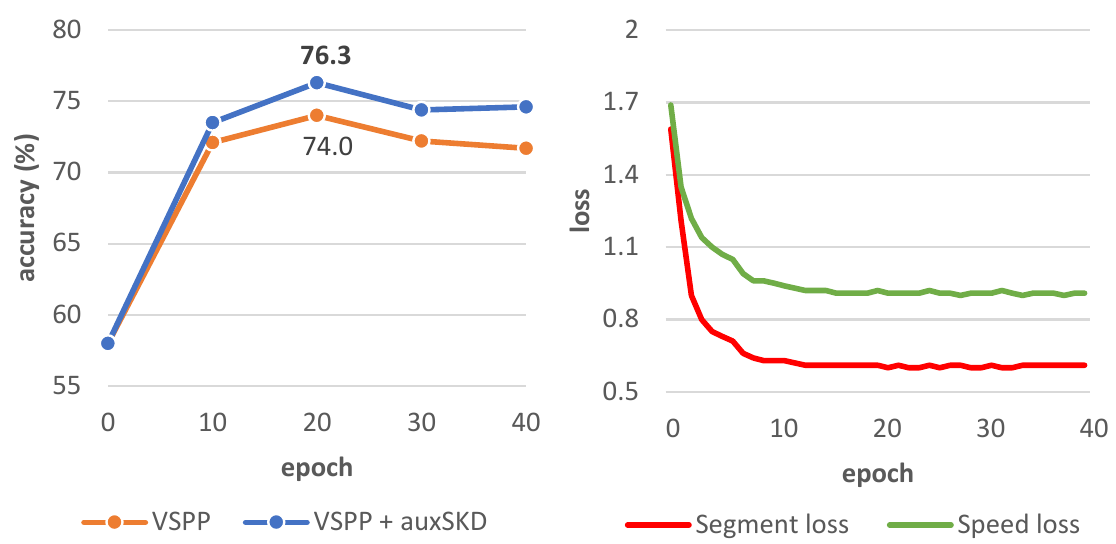}}
\caption{{(Left) VSPP pretext task performance with auxSKD (VSPP+auxSKD) and without (VSPP) based on the number of epochs. We pretrained the R(2+1)D model on K-100 for 40 epochs and report the results every 10 epochs on UCF101. (Right) Pre-training losses of our VSSP subtasks on K-100, i.e. speed prediction and segment prediction losses, further illustrate that our model actually converges after around 20 epochs.}}
\label{fig:curve}  
\end{figure}

{{\bf Ablation on VSPP} --  Our VSPP pretext task determines both the segment within a clip where there is a speed alteration compared to the natural speed of the rest of the clip and what the speed rate is, effectively parameters $\lambda$ and $\zeta$}.  Based on ablation studies in \cite{wang2020self}, for all the experiments here we consider 4 different speed rates i.e. $Q=4$, hence $\lambda = \{1, 2, 3, 4\}$.}}

{Table \ref{tab:ablateVSPP} outlines the effect of each sub-task in VSPP when our model pretrains on them separately and jointly (on K-100). Our auxSKD pretraining  is not engaged for this ablation. The best result is obtained at 60.8\% on the UCF101 dataset when pretraining jointly on both tasks and having the maximum number of segments $Z=4$ . }

{When the number of segments $Z$ during sampling is fewer (i.e. as $\zeta$ ranges from 1 to $Z$) or a subtask is missed out, the performance drops.  Note, the first line of the table when there is only one segment, i.e. $Z=1$ is the equivalent to VideoPace. We believe that increasing the number of segments in the clip pushes the model to temporally explore the video more to find that specific segment with different speed, resulting in better temporal representation. }

{\bf{Pretraining epochs}} --
\mm{ 
In Figure \ref{fig:curve} (left), we evaluate the performance of our method on UCF101 when pretrained on K-100, with and without auxSKD, using different checkpoints. It can be seen that when auxiliary learning is switched on, the performance of our VSPP pretext task is increased at all checkpoints. We also notice that the performance starts to saturate after 20 epochs for both VSPP and VSPP+auxSKD. In Figure \ref{fig:curve} (right), we can see that after around 20 epochs, the changes in the VSPP losses are not significant. This demonstrates that our model converges quickly and  we can ensure its convergence during pretraining with only 20 epochs.}}

\section{Limitations}

We identify three limitations of our work:


   (i) a fundamental aspect of our VSPP pretext task is that it thrives on  the altered natural pace of motion in a segment of a video while the rest of the clip retains its natural motion. However, any sudden and very fast motion in a clip may violate this assumption as the fast motion within the selected segment of a clip may be missed when it's sampled. This is a similar limitation for other current speed based pretext tasks such as VideoPace and RSPNet. 
    (ii) we aim to test our approach on at least one other reduced-size version of an existing, large dataset, such as FineAction \cite{FA} to further validate our auxiliary pretraining stage as an approach that reduces the dependence of self-supervised learning approaches on large pretraining datasets
    (iii) other speed-related pretext tasks, such as ASCNet, RSPNet, and VideoPace include an appearance stream in their methodology, however in this work, while the absence of an appearance stream may seem to be a limitation, it was avoided to focus on the power of VSPP as an independent pretext task and promote the auxiliary pretraining stage as two contributions that may be used in a modular fashion by the community.  We expect that adding an appearance stream to our \mm{model} may improve our results.

\section{Conclusions}
{In this paper, we introduced an auxiliary-learning phase for self-supervised video representation learning that allows a significant reduction in the amount of unlabelled data required for the pretraining task. The approach exploits similarity-based knowledge distillation to better prepare a (student) network to perform its primary pretraining task. Our experiments show that this new auxiliary phase auxSKD improves the performance of other existing SSL approaches, such as VCOP \cite{xu2019self}, VideoPace\cite{wang2020self}, and RSPNet \cite{chen2021rspnet}.} {We also introduced a new video speed analysis task, VSPP,  that predicts the index and altered speed of a segment within a clip which is sampled at a different {frame} rate to the rest of the clip. Solving this task can strength the network's awareness of the video's natural speed rate and alleviate  the imprecise video speed labeling problem \cite{chen2021rspnet}. Our experiments illustrate that the features learnt achieve competitive or superior results compared to the state of the art, while training on a much smaller dataset, e.g. K-100 rather than K-400, and at a lower computational cost.}


\section*{Acknowledgements}
The authors are sincerely grateful for the kind donations to the Southmead Hospital Charity and from Caroline Belcher. Their generosity has made this research possible.

{\small
\bibliographystyle{ieee_fullname}
\bibliography{egbib}
}

\end{document}